\documentclass{article}

\PassOptionsToPackage{numbers, compress}{natbib}

\usepackage[preprint]{nips_2018}
\usepackage{amsmath}
\usepackage{algorithm}
\usepackage[noend]{algpseudocode}

\makeatletter
\def\BState{\State\hskip-\ALG@thistlm}
\makeatother

\usepackage{multicol}

\usepackage[utf8]{inputenc} 
\usepackage[T1]{fontenc}    
\usepackage{hyperref}       
\usepackage{url}            
\usepackage{booktabs}       
\usepackage{amsfonts}       
\usepackage{nicefrac}       
\usepackage{microtype}      
\usepackage{graphicx}
\usepackage{multirow}

\title{Training Domain Specific Models for Energy-Efficient Object Detection} 

\author{
  Kentaro Yoshioka\\
  Department of Computer Science\\
  Stanford University, Keio University, Toshiba\\
  \texttt{kentaroy@stanford.edu} \\
   \And
  Edward Lee\\
  Department of Electronic Engineering\\
  Stanford University\\
  \texttt{edhlee@stanford.edu} \\
   \And
  Mark Horowitz\\
  Department of Computer Science\\
  Stanford University\\
  \texttt{horowitz@ee.stanford.edu} \\
}

\begin{document}

\maketitle

\begin{abstract}
We propose an end-to-end framework for training domain specific models (DSMs) to obtain both high accuracy and computational efficiency for object detection tasks. DSMs are trained with knowledge distillation  \cite{hinton2015distilling} and focus on achieving high accuracy at a limited domain (e.g. fixed view of an intersection or your office). We argue that DSMs can capture essential features well despite its small network capacity, enabling higher accuracy and efficiency.  In addition, since such training may be conducted directly on edge devices, we improve the training efficiency by reducing the dataset size by culling easy to classify images from the training set. For the limited domain, we observed that compact DSMs significantly surpass the accuracy of COCO trained models of the same size. By training on a compact dataset, we show that with an accuracy drop of only 3.6\%, the training time can be reduced by 93\%.
\end{abstract}

\section{Introduction}
Implementing CNN based object detection on stationary surveillance cameras can lead to enhancing the safety of cities, homes, offices and factories by detecting unauthorized substances or discovering anomaly events (e.g. a collapsed person). However, the computation-efficiency is a key requirement since such devices demand battery operation to ease installation. Many successful approaches to improve the efficiency of image classification have been proposed, such as model compression\cite{han2015deep} and model cascades with domain specific models \cite{kang2017noscope}.  However, object detection is more complex than image classification, and while these techniques are likely to remain effective, there is need for additional methods. 

Instead of compressing large models, we target to train a computation-efficient model for \textit{each specific} surveillance camera and a framework is proposed to train such domain specific models (DSM). The framework is based on knowledge distillation \cite{hinton2015distilling}\cite{chen2017learning}\cite{girshickdata} but targets to reduce the accuracy gap between student and teacher models by training the student using a restricted class of domain specific images. Since such training may be conducted on edge-devices, we improve the training efficiency by culling easy-to-classify images with small accuracy penalty.



This paper's contribution is summarized below.
\begin{itemize}
\item We propose an end-to-end framework for training domain specific models (DSMs) to mitigate the tradeoff between object-detection accuracy and computational efficiency. To the best of our knowledge, this is the first successful demonstration of training DSMs for object detection tasks. 
\item By training resnet18-based Faster-RCNN DSMs, we observed a 19.7\% accuracy (relative mAP) improvement compared to COCO trained model of the same size, tested on a customized YoutubeLive dataset.
\item Since edge devices will have limited resources, we propose culling the training dataset to significantly reduce the computation resource required for the training. Only training data that has high utility in training is added. This filtering allows us to reduce training time by 93\% with an accuracy loss of only 3.6\%.
\end{itemize}

\section{Training Domain Specific Models}
Large scale object detection datasets such as COCO\cite{lin2014microsoft} contain a large and diverse set of natural images. Using a small model on such a large dataset would typically yield higher misdetections than a large model. Furthermore, \cite{chen2017learning} showed that misdetections usually occur between foreground and background (false positives $+$ true negatives); rarely do misdetections occur as a result of inter-class errors. In video surveillance, because frames in a video stream share a stationary background, a compact model can be good enough to detect foreground and background. This motivates our DSM framework to train compact models with dataset constructed by domain-specific data.

\begin{algorithm}[!t]
\caption{Training Domain Specific Models}\label{euclid}
\begin{algorithmic}[1]
\Require Domain Specific Model (DSM), Teacher Model
\Procedure{1. Prepare Dataset}{}

Given domain images $x_i$ for $i = \{0, \dots, N_{\mathrm{train}}-1 \} $   \;
 \For{$i<N_{\mathrm{train}}$}
 \State $\mathrm{label}(i) \leftarrow \mathrm{Teacher\texttt{.predict}}(x_i)$. 
 \State $\mathrm{pred}(i) \leftarrow \mathrm{DSM\texttt{.predict}}(x_i)$.
 \State compute  $L_{\mathrm{train}}(i)$ from $\mathrm{label}(i)$ and $\mathrm{pred}(i)$.
 \EndFor
 
 \State Collect $[x_i, \mathrm{label}(i)]$ pairs with $n$ largest values of $L_{\mathrm{train}}$.
 \BState \emph{Compile Difficult Dataset (DDS)}:
 $\Omega = ( [x_0, \mathrm{label}(0)]), \dots, [(x_{n-1}, \mathrm{label}(n-1)])$.
\EndProcedure

\Procedure{2. Train DSM}{}
 \State DSM\texttt{.train} $(\Omega)$
\EndProcedure

\Procedure{3. Inference}{}
\State Detection $\leftarrow$ DSM\texttt{.predict}(image)
\EndProcedure

\end{algorithmic}
\end{algorithm}

\begin{figure}[!t]
\centering
  \includegraphics[width=11cm, bb=150 0 1000 180]{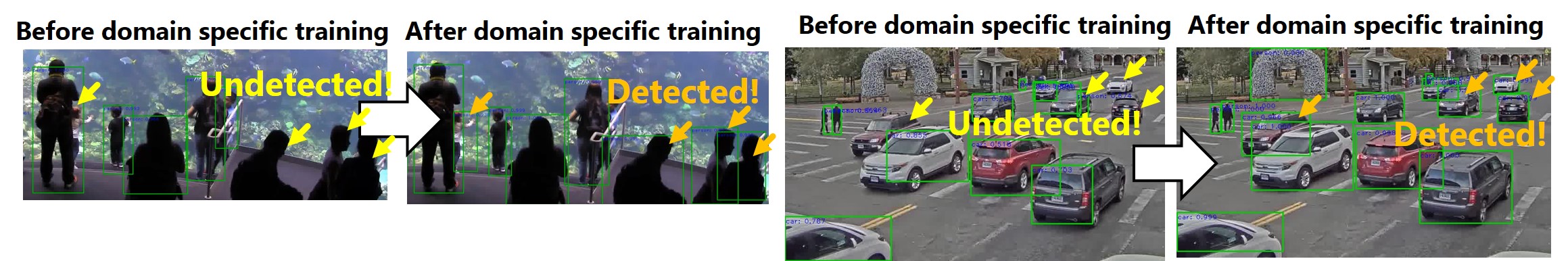}
  \caption{Object detection results of the test image, before and after domain specific training.}
\end{figure}

As illustrated in Algorithm 1, our DSM framework consists of preparation of the data and training of the DSM. A large challenge when deploying models in surveillance is preparing the training data since manually labelling frames in videos is cumbersome. 
To overcome this, we label the dataset used to train the DSM by using the predictions of a much larger teacher model with higher accuracy and treating these predictions as ground truth labels. Furthermore, we compare the prediction on image $x_i$ made by the teacher to that of the DSM; we determine whether to store the $x_i$ and label Teacher$.\texttt{predict}(x_i)$ in our compiled dataset $\Omega$. After the training set is compiled, it is used to train the DSM.

Training a object detection model can take hours even with a GPU and can be challenging for applications requiring frequent retraining.
We exploit the fact that when the DSM is pretrained on large-scale general dataset, it can already provide good predictions for a large chunk of the domain-specific data. This procedure develops a compact dataset $\Omega$ that is only composed of data that the DSM finds inconsistent with the prediction made by the teacher model. Keeping data $x_j$ that both the DSM and teacher detections are consistent is computationally redundant because it does not contribute to gradient signals. We define $L_{train}$ to quantify the consistency between teacher and DSM:
\begin{eqnarray}
\centering
L_{train}  & = & \frac{FP + TP}{TP + \epsilon} + \frac{FN + TP}{TP + \epsilon} 
\end{eqnarray}
where TP, FP, FN represents the number of true positive, false positive and false negative bounding-box (BB) detections of the image and $\epsilon = 0.5$. Significantly fewer training data and steps are required with only a minimal penalty in accuracy.

\begin{table}[]
\centering
  \caption{ Domain specific training results are summarized, where the mean accuracy result of 5 datasets are reported. Res101 results are used as ground truth, therefore accuracy is relative mAP (rmAP). Along the model name, parameters and inference time on GPU per image is reported.}
\begin{tabular}{|c|l|c|c|c|c|c|c|}
\hline
\multicolumn{2}{|c|}{\multirow{3}{*}{}} & \multicolumn{6}{c|}{Teacher: Res101 {[}48M/68ms{]}}                                   \\ \cline{3-8} 
\multicolumn{2}{|c|}{}                  & \multicolumn{3}{c|}{Res18 {[}12M/26ms\}} & \multicolumn{3}{c|}{Squeeze {[}6M/21ms{]}} \\ \cline{3-8} 
\multicolumn{2}{|c|}{}                  & COCO     & DSM      & Improvement        & COCO      & DSM      & Improvement         \\ \hline
\multicolumn{2}{|c|}{mean accuracy}           & 54.5     & 74.3     & \textbf{+ 19.7}    & 41.5      & 63.3     & \textbf{+ 21.7}     \\ \hline
\end{tabular}

  \label{table:dsm results}
\end{table}

\begin{table}[]
\centering
\caption{Number of training samples versus accuracy with res18. For simple, we pick the first N training data and filter out the rest. For difficult dataset (DDS), N training data having highest $L_{train}$ are chosen. The mean accuracy drop of the 5 datasets were computed, in respect to the model trained with all 3600 images. The training time does not include the time for teacher model labeling, which takes about 10 min. We utilize single TitanXp GPU to measure the training time.}
\begin{tabular}{|c|c|c|c|c|c|c|c|}
\hline
\multirow{2}{*}{Dataset}                                     & \multirow{2}{*}{Classes} & \multirow{2}{*}{Strategy}                            & \multicolumn{5}{c|}{Number of training samples $n$}                                                                                                                                                                                                                            \\ \cline{4-8} 
                                                             &                          &                                                      & 64                                                  & 128                                                 & 256                                                 & 512                                                 & \begin{tabular}[c]{@{}c@{}}All\\ (3600)\end{tabular} \\ \hline
coral                                                        & 1                        & \begin{tabular}[c]{@{}c@{}}Simple\\ DDS\end{tabular} & \begin{tabular}[c]{@{}c@{}}81\\ 89.6\end{tabular}   & \begin{tabular}[c]{@{}c@{}}89.4\\ 89.6\end{tabular} & \begin{tabular}[c]{@{}c@{}}89.6\\ 89.6\end{tabular} & \begin{tabular}[c]{@{}c@{}}90\\ 89.8\end{tabular}   & 90                                                   \\ \hline
taipei                                                       & 4                        & \begin{tabular}[c]{@{}c@{}}Simple\\ DDS\end{tabular} & \begin{tabular}[c]{@{}c@{}}50.4\\ 60.7\end{tabular} & \begin{tabular}[c]{@{}c@{}}62\\ 61.7\end{tabular}   & \begin{tabular}[c]{@{}c@{}}62.1\\ 62.2\end{tabular} & \begin{tabular}[c]{@{}c@{}}62.8\\ 64.2\end{tabular} & 68.2                                                 \\ \hline
jackson                                                      & 2                        & \begin{tabular}[c]{@{}c@{}}Simple\\ DDS\end{tabular} & \begin{tabular}[c]{@{}c@{}}52.5\\ 71.6\end{tabular} & \begin{tabular}[c]{@{}c@{}}60.1\\ 76.7\end{tabular} & \begin{tabular}[c]{@{}c@{}}60.9\\ 78.3\end{tabular} & \begin{tabular}[c]{@{}c@{}}72.8\\ 80.6\end{tabular} & 87.0                                                 \\ \hline
kentucky                                                     & 2                        & \begin{tabular}[c]{@{}c@{}}Simple\\ DDS\end{tabular} & \begin{tabular}[c]{@{}c@{}}35\\ 53.1\end{tabular}   & \begin{tabular}[c]{@{}c@{}}38.7\\ 63.8\end{tabular} & \begin{tabular}[c]{@{}c@{}}44.2\\ 66.4\end{tabular} & \begin{tabular}[c]{@{}c@{}}54.8\\ 69.5\end{tabular} & 67.2                                                 \\ \hline
castro                                                       & 3                        & \begin{tabular}[c]{@{}c@{}}Simple\\ DDS\end{tabular} & \begin{tabular}[c]{@{}c@{}}60.4\\ 67.2\end{tabular} & \begin{tabular}[c]{@{}c@{}}63\\ 68.35\end{tabular}  & \begin{tabular}[c]{@{}c@{}}65.2\\ 75.0\end{tabular} & \begin{tabular}[c]{@{}c@{}}66\\ 77\end{tabular}     & 77.6                                                 \\ \hline
\begin{tabular}[c]{@{}c@{}}mean\\ accuracy drop\end{tabular} & -                        & \begin{tabular}[c]{@{}c@{}}Simple\\ DDS\end{tabular} & \begin{tabular}[c]{@{}c@{}}23.6\\ \textbf{9.5}\end{tabular}  & \begin{tabular}[c]{@{}c@{}}20.8\\ \textbf{5.9}\end{tabular}  & \begin{tabular}[c]{@{}c@{}}13.8\\ \textbf{3.6}\end{tabular}  & \begin{tabular}[c]{@{}c@{}}11.3\\ \textbf{1.7}\end{tabular}  & 0                                                    \\ \hline
Training Time  {[}min{]}                                     & -                        & -                                                    & \textbf{1.8}                                        & \textbf{3.6}                                        & \textbf{7.4}                                        & \textbf{14.6}                                       & 110                                                  \\ \hline
\end{tabular}

\label{table:DDS table}
\end{table}

\section{Experiments}
\textbf{Models. } 
We develop Faster-RCNN object detection models on PyTorch pretrained on MS-COCO\cite{paszke2017pytorch}\cite{ren2015faster}\cite{lin2014microsoft}. We use 3 models:  resnet101(res101), resnet(res18), and squeezenet(squeeze) as the backbone region proposal network (RPN) \cite{he2016deep}\cite{iandola2016squeezenet}. Res18 and squeeze holds 10\% and 19\% TOP-5 Imagenet error, which is a common accuracy range for compact CNN models like MobileNet \cite{howard2017mobilenets}. During training, res101 is used as the teacher, while res18 and squeeze are used as DSMs. While we chose Faster-RCNN for its accuracy on YoutubeLive, we can also use YOLO/SSD detectors for improved efficiency with this framework because the training requires the bounding box labels \cite{redmon2016you}\cite{liu2016ssd}.
\footnote{We release the codes and the dataset 

https://github.com/kentaroy47/training-domain-specific-models}

\textbf{Dataset. } 
We obtain 5 fixed-angle videos from \texttt{YouTubeLive}.
The video is 2 hours each with 1 frames-per-second (fps), consisting of 7200 images. We split the images evenly: the first 3600 images are for training and the later 3600 images for testing. 

\textbf{Results. }
As shown in table~\ref{table:dsm results}, we first train our res18 DSM using the full $N_{\mathrm{train}}=$ 3600 training images for 10 epochs using stochastic-gradient descent with a learn rate of $10^{-4}$. As compared to the res18 model pretrained on MS-COCO but without domain specific training, we achieved an average of 19.7\% accuracy improvement.


Table 2 shows the effectiveness of DDS on res18. Using DDS, we were able to reduce the training time by 93\% (256 images) with only 3.6\% accuracy penalty. If we simply picked 256 training images sequentially (strategy simple on table), the accuracy worsens 10.2\% compared to DDS. 

\subsection{Comparison against Data Distillation}
Data Distillation \cite{girshickdata} is fundamentally different from our application setting and methods. Models with large network capacities were shown to achieve higher accuracy by bootstrapping the dataset with  \cite{girshickdata}. On the other hand in our framework, in order to fully utilize the small network capacity, we aim to train the models with only the domain specific data.

We show on Table 3 that following a method of data distillation (i.e. aggregating PASCAL with the entire YoutubeLive data) yields lower rmAP improvement than with our approach of training with only domain specific data. In addition, training the small models with the entire YoutubeLive dataset also yields lower improvements as well. This is fundamentally because of the limited model capacity of the compact, but computationally-efficient model. We observe that for training small models, utilizing a larger dataset do not always obtain better results but restricting the data domain can do better.  

    \begin{table}[!t]
    \centering
    \caption{Accuracy improvement observed for multiple dataset settings. For PASCAL+YoutubeLive, we train the models with PASCAL-VOC2007 and YoutubeLive data. The accuracy improvement is rmAP improvement compared to the COCO trained model.}
    \begin{tabular}{|c|l|c|c|c|}
    \hline
    \multicolumn{2}{|c|}{}                                                                            & PASCAL+YoutubeLive & YoutubeLive & Domain Specific \\ \hline
    \multicolumn{2}{|c|}{\begin{tabular}[c]{@{}c@{}}mean accuracy improvement\\ Res18\end{tabular}}   & + 9.7              & + 12.9      & + 19.7          \\ \hline
    \multicolumn{2}{|c|}{\begin{tabular}[c]{@{}c@{}}mean accuracy improvement\\ Squeeze\end{tabular}} & + 10.3             & + 13.5      & + 21.7          \\ \hline
    \end{tabular}
    \end{table}

\small
\bibliographystyle{unsrt}
\bibliography{main}

\normalsize

\end{document}